\documentclass[11pt]{article}

\usepackage[preprint]{acl}

\usepackage{times}
\usepackage{latexsym}

\usepackage[T1]{fontenc}

\usepackage[utf8]{inputenc}

\usepackage{microtype}
\usepackage{amsmath}
\usepackage{amssymb}
\usepackage{inconsolata}

\usepackage{graphicx}

\usepackage{booktabs}
\usepackage{multirow}
\usepackage{makecell}
\usepackage{graphicx} 
\usepackage{colortbl}
\usepackage{xcolor}

\title{Exons-Detect: Identifying and Amplifying Exonic Tokens via Hidden-State Discrepancy for Robust AI-Generated Text Detection}

\author{Xiaowei Zhu\textsuperscript{1,2}, Yubing Ren\textsuperscript{1,2}\thanks{Corresponding author.}, Fang Fang\textsuperscript{1,2}, \\ \textbf{Shi Wang\textsuperscript{3}, Yanan Cao\textsuperscript{1,2}, Li Guo\textsuperscript{1,2}}\\
        \textsuperscript{1}Institute of Information Engineering, Chinese Academy of Sciences, Beijing, China \\ 
        \textsuperscript{2}School of Cyber Security, University of Chinese Academy of Sciences, Beijing, China \\
        \textsuperscript{3}	Institute of Computing Science, Chinese Academy of Sciences, Beijing, China \\
        \texttt{\{zhuxiaowei, renyubing\}@iie.ac.cn}}

\begin{document}
\maketitle
\begin{abstract}
The rapid advancement of large language models has increasingly blurred the boundary between human-written and AI-generated text, raising societal risks such as misinformation dissemination, authorship ambiguity, and threats to intellectual property rights. These concerns highlight the urgent need for effective and reliable detection methods. While existing training-free approaches often achieve strong performance by aggregating token-level signals into a global score, they typically assume uniform token contributions, making them less robust under short sequences or localized token modifications. To address these limitations, we propose Exons-Detect, a training-free method for AI-generated text detection based on an exon-aware token reweighting perspective. Exons-Detect identifies and amplifies informative exonic tokens by measuring hidden-state discrepancy under a dual-model setting, and computes an interpretable translation score from the resulting importance-weighted token sequence. Empirical evaluations demonstrate that Exons-Detect achieves state-of-the-art detection performance and exhibits strong robustness to adversarial attacks and varying input lengths. In particular, it attains a 2.2\% relative improvement in average AUROC over the strongest prior baseline on DetectRL.
\end{abstract}

\section{Introduction}

The rapid advancement of large language models (LLMs) has enabled them to generate highly fluent and coherent text, substantially narrowing the observable gap between AI-generated text and human writing. While such progress has catalyzed significant technological breakthroughs across both industry and academia, it has simultaneously introduced pressing societal risks, including misinformation dissemination, authorship ambiguity, and threats to intellectual property rights \citep{ahmed2021detectingfakenewsusing, Adelani2019GeneratingSF, 9121286}. Prior studies \cite{clark-etal-2021-thats} further reveal that humans perform only marginally above random chance in distinguishing AI-generated from human-written text. This limitation highlights the urgent need for effective and reliable detection methods.

\begin{figure}[t]
\centering
\counterwithout{figure}{section}
\resizebox{\linewidth}{!}{
\includegraphics{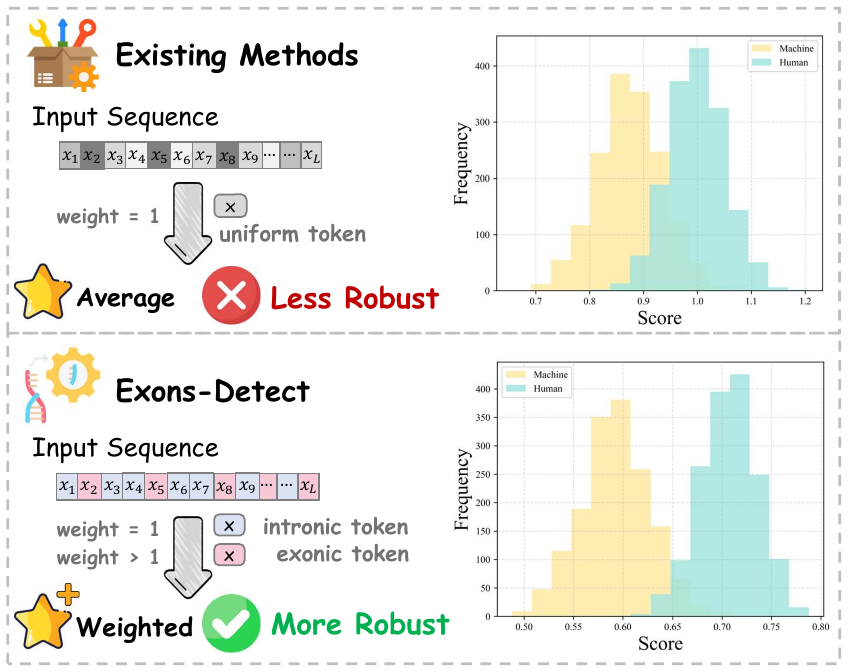}
}
\caption{Advantages of Our Method Exons-Detect.}
\label{fig:intro}
\end{figure}

Existing detection approaches can be broadly categorized into training-based and training-free methods. Training-based methods require large-scale labeled data and rely on supervised deep models to learn implicit textual representations, which limits their scalability and cross-domain generalization. In contrast, training-free methods compute token-level statistics, such as Binoculars \cite{hans2024spotting}, under the generative distributions of proxy LLMs, and typically aggregate these signals by averaging across token positions to form a global detection score. While effective in many settings, this uniform aggregation assumes equal contribution from all tokens, making such methods less robust when token sequences are short, or localized token modifications are introduced. Consequently, truly informative tokens can be overwhelmed by less relevant ones, motivating the need for a mechanism that differentiates token-level functional roles rather than treating all tokens uniformly.

Inspired by molecular biology, we view a text sequence as a gene fragment composed of exons and introns. Exons are directly involved in protein translation, while introns play a secondary role. Analogously, tokens with large hidden-state discrepancy are treated as exonic tokens that carry stronger discriminative signals, whereas the remaining tokens are regarded as intronic. During transcription, both exons and introns are preserved in the pre-mRNA, corresponding to uniform initial weighting of all tokens. Splicing then removes introns and emphasizes exons in mature mRNA. Mirroring this process, we identify exonic tokens via hidden-state discrepancy and amplify their contributions through assigning additional weights. The final detection is achieved by aggregating the reweighted token sequence to compute the translation score. This exon-aware reweighting captures intrinsic differences between AI-generated and human-written texts in a fine-grained and interpretable manner.

Building on this intuition, we propose Exons-Detect, a novel training-free method for AI-generated text detection. Given an input sequence, we extract hidden representations at each token position and quantify their discrepancy under a pair of proxy LLMs. Tokens whose representation discrepancy exceeds a predefined discrepancy threshold are identified as exonic tokens. We map these discrepancies through a nonlinear function to obtain additional weights for exonic tokens, which are integrated with the initial weights for computing the translation score. Finally, Exons-Detect determine the detect result by comparing the translation score against a decision threshold, effectively amplifying the discriminative signals carried by exonic tokens.

Exons-Detect achieves state-of-the-art performance across multiple publicly available detection benchmarks. In particular, Exons-Detect achieves a relative improvement of 2.2\% in average AUROC over the strongest existing baseline DNA-DetectLLM on DetectRL.  Moreover, Exons-Detect exhibits strong robustness against various adversarial attacks and across different input lengths. Efficiency experiments further demonstrate that Exons-Detect offers rapid detection capability, making it well suited for large-scale, real-time detection. Our contributions are summarized as follows:
\begin{itemize}
\item Inspired by the distinct roles of exons and introns in gene fragments, we introduce the notions of exonic tokens and intronic tokens in text sequences, emphasizing that different tokens contribute unequally to detection.
\item We propose Exons-Detect, a novel training-free method that identifies exonic tokens and amplifies their importance to capture more informative source-specific signals, enabling robust AI-generated text detection.
\item Extensive experiments demonstrate that Exons-Detect provides a robust, efficient, and broadly generalizable solution for AI-generated text detection, delivering consistent improvements across 3 public benchmarks, 2 adversarial attacks, and varying input lengths, with inference latency below 0.8 s per sample.
\end{itemize}

\begin{figure*}[t]
\centering
\counterwithout{figure}{section}
\resizebox{\linewidth}{!}{
\includegraphics{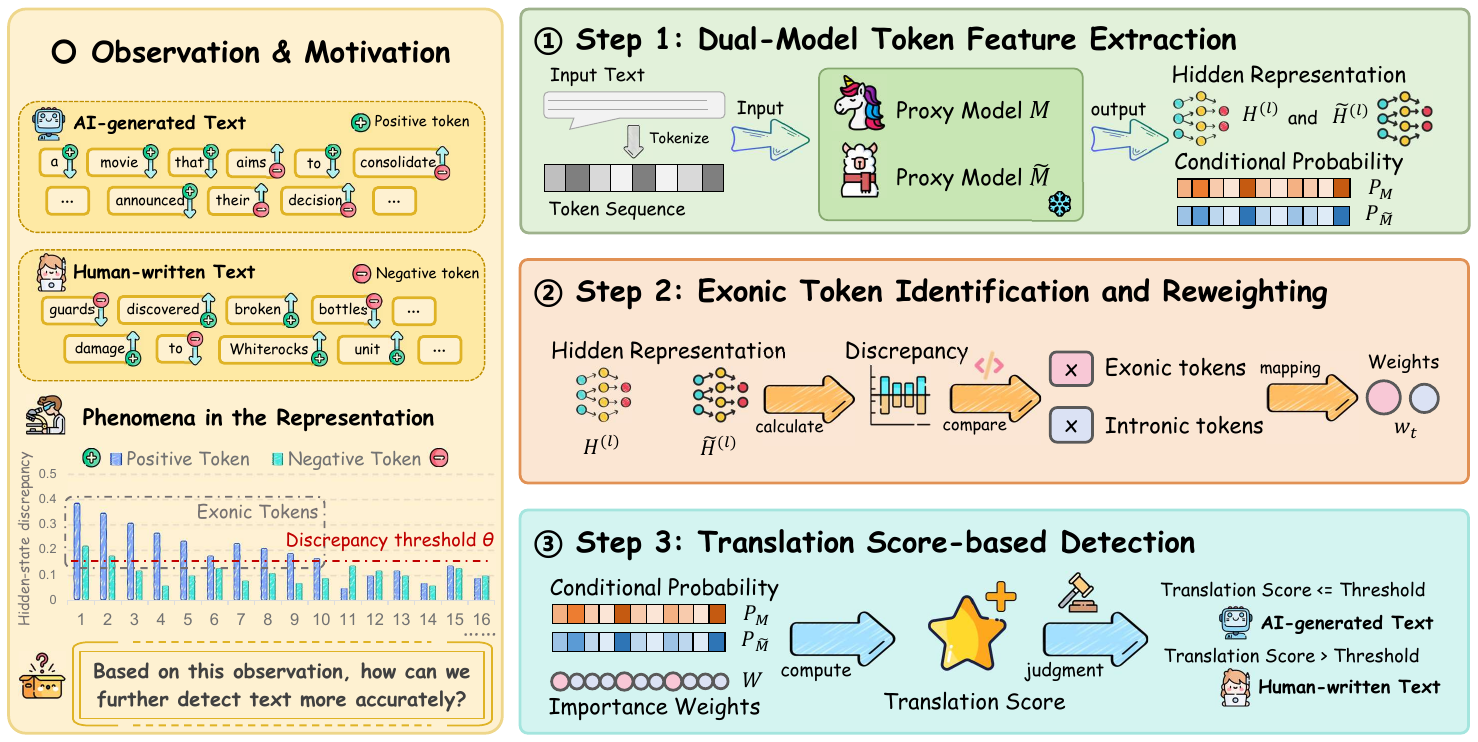}
}
\caption{Overview of Exons-Detect.}
\label{fig:method}
\end{figure*}

\section{Related Work}

\noindent\textbf{Training-based methods} typically leverage deep learning models to supervisedly learn latent textual features that distinguish AI-generated text from human-written content. Early work by OpenAI \cite{solaiman2019release} developed a RoBERTa-based classifier. RADAR \cite{NEURIPS2023_30e15e59} incorporated adversarial training to improve robustness against paraphrased inputs. Biscope \cite{NEURIPS2024_bc808cf2} introduced a bidirectional cross-entropy loss to optimize classifier performance. DeTeCtive \cite{NEURIPS2024_a117a3cd} and DETree \cite{he2025detree} mapped texts from different sources or constructions into high-dimensional representation spaces, followed by similarity-based detection. Training-based detectors often overfit in-distribution patterns and degrade sharply under distribution shifts~\cite{chakraborty2023possibilitiesaigeneratedtextdetection, uchendu-etal-2020-authorship}, motivating increasing interest in universal and reliable training-free detection.

\noindent\textbf{Training-free methods} distinguish texts by estimating statistical scores from the generative probabilities under proxy LLMs. Traditional approaches including LogRank \cite{gehrmann-etal-2019-gltr}, Likelihood \cite{hashimoto-etal-2019-unifying}, and Entropy \cite{ippolito-etal-2020-automatic} quantified generative uncertainty by averaging probability rank, log-likelihood, and entropy under a proxy model. DetectGPT~\cite{pmlr-v202-mitchell23a} established a new paradigm by introducing phrase-level perturbations to evaluate distributional curvature. Fast-DetectGPT~\cite{bao2024fastdetectgpt} proposed an optimized sampling strategy for estimating conditional probability curvature, achieving substantial gains in both speed and accuracy compared to DetectGPT. Binoculars~\cite{hans2024spotting} mitigated high-perplexity human texts by using the ratio of log-perplexity to cross-perplexity, while DNA-DetectLLM~\cite{zhu2025dnadetectllm} further enhanced this score with a mutation-repair mechanism, achieving more robust performance. Lastde~\cite{DBLP:journals/corr/abs-2410-06072} captured local textual characteristics via Diversity Entropy, while IRM~\cite{liu2025zeroshot} leveraged discrepancies in generative probabilities before and after reinforcement learning from human feedback (RLHF) to capture the divergence.

\section{Methodology}

\subsection{Preliminaries}
\paragraph{Log-perplexity and Cross-perplexity.}
Log-perplexity quantifies the average token-level negative log-likelihood under a single proxy model, whereas cross-perplexity measures the average per-token cross-entropy computed across two models. To model variation in token-wise importance, we further introduce weighted log-perplexity and weighted cross-perplexity, which incorporate token-specific importance weights into their computation:
\begin{equation}
{\footnotesize
\begin{aligned}
\log \mathrm{PPL}^W_{M}(s)
&= - \sum_{t=1}^{T} w_t \log P_{M}(x_t), \\
\log \mathrm{X\text{-}PPL}^W_{M,\tilde{M}}(s)
&= - \sum_{t=1}^{T} w_t \,
   P_{M}(x_t) \,
   \log P_{\tilde{M}}(x_t),
\end{aligned}
}
\label{eq:ppl-xppl}
\end{equation}

\noindent
where $s$ denotes an input sequence of length $T$, $w_t$ denotes the normalized weight, and $P_M(x_t)$ and $P_{\tilde{M}}(x_t)$ denote the conditional generation distributions of the $t$-th token under models $M$ and $\tilde{M}$.


\paragraph{Observation.}
In training-free detection, scores such as Binoculars are computed by averaging token-level contributions, where AI-generated texts typically yield lower scores than human-written texts. Accordingly, in AI-generated text, tokens whose individual contributions fall below the decision threshold tend to increase class separability, whereas in human-written text, tokens with contributions above the threshold play an analogous role. We refer to such tokens as positive tokens and, by analyzing their associated hidden-state representations that may encode source-related signals \cite{chen2025repreguarddetectingllmgeneratedtext}, as shown in Figure~\ref{fig:method}.
When we examine tokens whose hidden-state discrepancy between model $M$ and $\tilde{M}$ exceeds a discrepancy threshold $\theta$, the number of tokens that increase class separability is markedly larger than the number that decreases it. This asymmetric enrichment suggests that tokens with high-discrepancy more often carry source-relevant signals, whereas other tokens contribute more weakly or inconsistently.

\paragraph{Motivation.}
Motivated by this observation, we refer to tokens with high hidden-state discrepancies as \textbf{exonic tokens}, and the remaining ones as \textbf{intronic tokens}, reflecting their different relevance to the text’s origin. This naturally suggests a simple and effective strategy: during detection, we identify high-discrepancy exonic tokens and amplify their contributions. By reweighting these tokens, the final detection score is encouraged to move further toward the correct side, resulting in robust and separable detection. 
See Appendix~\ref{app:exon-proof} for analysis.

\subsection{Overview of Exons-Detect}
Figure~\ref{fig:method} presents the overall workflow of Exons-Detect, including three steps:

\noindent\textbf{Step 1: Dual-Model Token Feature Extraction.}
Given an input sequence, we extract token-level hidden representations and generative probability distributions under a reference model $M$ and a paired model $\tilde{M}$.

\noindent\textbf{Step 2: Exonic Token Identification and Reweighting.} 
We measure hidden-state discrepancy between $M$ and $\tilde{M}$ at each token position to identify exonic tokens, and map these discrepancies to additional token-level weights.

\noindent\textbf{Step 3: Translation Score-based Detection.}
We introduce a translation score by aggregating token contributions according to their weights and probability distributions, and compare it against a decision threshold to determine the detection result.

\subsection{Dual-Model Token Feature Extraction}
Given an input sequence $s = (x_1, x_2, x_3, \dots, x_T)$ of length $T$, we feed it into a proxy LLMs pair: $M$ and $\tilde{M}$. Each model consists of $L$ transformer layers. For model $M$, we extract the hidden representations at each token position and layer as
\begin{equation}
\mathbf{H}^{(l)} =
\big[
\mathbf{h}^{(l)}_1,
\mathbf{h}^{(l)}_2,
\dots,
\mathbf{h}^{(l)}_T
\big],
\quad l = 1, \dots, L,
\end{equation}
where $\mathbf{h}^{(l)}_t \in \mathbb{R}^d$ denotes the hidden representation of the $t$-th token at layer $l$. Similarly, model $\tilde{M}$ produces the corresponding hidden representations
\begin{equation}
\tilde{\mathbf{H}}^{(l)} =
\big[
\tilde{\mathbf{h}}^{(l)}_1,
\tilde{\mathbf{h}}^{(l)}_2,
\dots,
\tilde{\mathbf{h}}^{(l)}_T
\big].
\end{equation}

In addition to hidden representations, both models provide token-level generative probabilities. Specifically, at each token position $t$, model $M$ defines a conditional generation distribution $P_M(x_t)=P_M(\cdot \mid x_{<t})$ over the vocabulary, and $\tilde{M}$ defines $P_{\tilde{M}}(x_t) = P_{\tilde{M}}(\cdot \mid x_{<t})$, where $x_{<t}$ denotes the preceding context.

\subsection{Exonic Token Identification and Reweighting}
To quantify the representational discrepancy at each token position, we measure the hidden-state discrepancy using the cosine distance. For the $t$-th token, we aggregate its hidden representations across all $L$ layers from models $M$ and $\tilde{M}$, and define the token-level discrepancy as
\begin{equation}
\delta_t
=
\frac{1}{L}
\sum_{l=1}^{L}
\left(
1 - \cos\big(
\mathbf{h}^{(l)}_t,\;
\tilde{\mathbf{h}}^{(l)}_t
\big)
\right),
\end{equation}
where $\mathbf{h}^{(l)}_t$ and $\tilde{\mathbf{h}}^{(l)}_t$ denote the hidden representations of the $t$-th token at layer $l$ produced by models $M$ and $\tilde{M}$, respectively.

Based on the magnitude of $\delta_t$, we identify exonic tokens by applying a significance-level criterion. Specifically, a token is classified as an exonic token if its hidden-state discrepancy exceeds a predefined discrepancy threshold $\theta$, and as an intronic token otherwise:
\begin{equation}
x_t =
\begin{cases}
\text{exonic token}, & \text{if } \delta_t > \theta, \\
\text{intronic token}, & \text{if } \delta_t \le \theta.
\end{cases}
\end{equation}

To further emphasize the contribution of exonic tokens, we remap their hidden-state discrepancies into importance weights $W$.  Formally, we introduce a nonlinear mapping function $g(\cdot)$ to obtain a token-specific additional weight $\Delta w_t = g(\delta_t)$:
\begin{equation}
g(\delta_t)
=
1 - \exp\!\big(-\alpha \, (\delta_t - \theta)_+\big),
\label{Eq:mapping}
\end{equation}
where  $(\cdot)_+ = \max(\cdot, 0)$ denotes the positive part operator, which ensures that intronic tokens with discrepancies below the discrepancy threshold $\theta$ receive zero additional weight.

This nonlinear mapping smoothly amplifies the weights of exonic tokens according to their hidden-state discrepancies, while avoiding excessive emphasis on individual tokens, thereby preserving robustness. The final importance weights is formed by summing the initial uniform weight and the exonic weight increments and normalizing the result, given by:
\begin{equation}
w_t = \frac{1 + \Delta w_t}{\sum_{i=1}^{T} \left(1 + \Delta w_i\right)}, 
\quad t = 1,2,\ldots,T.
\end{equation}

\subsection{Translation Score-based Detection}
We introduce a translation score that integrates the importance weights of both exonic and intronic tokens with their conditional probabilities. Prior work \cite{hans2024spotting} has shown that the ratio between log-perplexity and cross-perplexity provides a strong discriminative signal for AI-generated text detection. Following this insight, we define the initial translation score as the ratio of the weighted log-perplexity to the weighted cross-perplexity:
\begin{equation}
R(s) = \frac{\log \mathrm{PPL}^W_{M}(s)}{\log \mathrm{X\text{-}PPL}^W_{M,\tilde{M}}(s)}.
\end{equation}
To further refine the translation score, we incorporate the mutation-repair mechanism proposed in DNA-DetectLLM~\cite{zhu2025dnadetectllm} as a complementary component. This mechanism captures the intrinsic discrepancy between an input sequence and the ideal AI-generated sequence by quantifying the difficulty of iteratively repairing mutated tokens. Importantly, the repair process operates under the same exon-aware importance weights. Incorporating this mechanism, the final translation score is formulated as:

\begin{equation}
R(s) = \frac{\log \mathrm{PPL}^W_{M}(s) + \log \mathrm{PPL}^W_{M}(\hat{s})}{\log \mathrm{X\text{-}PPL}^W_{M,\tilde{M}}(s)},
\end{equation} 
where $\hat{s}$ denotes the ideal AI sequence, constructed by selecting the token with the maximum generation probability.

For AI-generated text, exonic tokens contribute to shifting the translation score toward smaller values, whereas for human-written text, exonic tokens contribute to increasing the translation score. Accordingly, the detection result for an input sequence is determined as follows:
\begin{equation}
    \mathcal{D}(s) = \begin{cases}\text { Human-written Text, } & R(s) > \tau, \\ \text { AI-generated Text, } & R(s)\leq\tau .\end{cases}
\end{equation}







\section{Experiments}

\subsection{Experimental Setup}
\paragraph{Datasets.}
To evaluate the detection performance of our method under realistic deployment scenarios, we conduct experiments on three diverse and high-quality public benchmarks: M4 \cite{wang-etal-2024-m4}, RealDet \cite{zhu-etal-2025-reliably}, and DetectRL \cite{wu2024detectrl}. In particular, we conduct evaluations on DetectRL using the Multi-LLM and Multi-Domain settings to examine generalization across models and domains.

\paragraph{Baselines.}
For training-based detectors, we consider OpenAI-D \cite{solaiman2019release}, BiScope \cite{NEURIPS2024_bc808cf2}, and R-Detect \cite{song2025deep}. For training-free approaches, we include classical zero-shot detectors such as Likelihood \cite{hashimoto-etal-2019-unifying}, LogRank \cite{gehrmann-etal-2019-gltr}, and Entropy \cite{ippolito-etal-2020-automatic}, as well as more recent representative methods, including DetectGPT \cite{pmlr-v202-mitchell23a}, Fast-DetectGPT \cite{bao2024fastdetectgpt}, Binoculars \cite{hans2024spotting}, and Lastde++ \cite{DBLP:journals/corr/abs-2410-06072}. In addition, we compare against the latest and strongest baselines, IRM \cite{liu2025zeroshot} and DNA-DetectLLM \cite{zhu2025dnadetectllm}.

\paragraph{Metrics.}
We evaluate detection performance using the area under the receiver operating characteristic curve (AUROC) and the F1 score.

\paragraph{Implementation details.}
To ensure a fair comparison, we train all training-based detectors on the HC3 dataset \cite{guo2023closechatgpthumanexperts}, which is disjoint from all evaluation benchmarks. Prior work \cite{bao2025glimpse} has shown that the performance of training-free methods can vary substantially under different combinations of LLMs. To eliminate this factor, we standardize the reference (scoring) model across all methods by using Falcon-7B-Instruct \cite{refinedweb} to compute token-level generation probabilities. Moreover, Fast-DetectGPT, Binoculars, Lastde++, IRM, DNA-DetectLLM, and Exons-Detect all employ Falcon-7B \cite{refinedweb} as the paired model when computing their respective detection scores. We set the discrepancy threshold $\theta=0.15$ and the mapping slope $\alpha=10$ by default. More details see Appendix~\ref{appendix:implementation}.

\begin{table*}[t]
\centering
\resizebox{\textwidth}{!}{
\begin{tabular}{l|cc|cc|cc|cc|cc}
\toprule
\multirow{2}{*}{\textbf{Detectors}} 
& \multicolumn{2}{c|}{\textbf{M4}} 
& \multicolumn{2}{c|}{\makecell{\textbf{DetectRL}  \textbf{Multi-LLM}}} 
& \multicolumn{2}{c|}{\makecell{\textbf{DetectRL}  \textbf{Multi-Domain}}} 
& \multicolumn{2}{c|}{\textbf{RealDet}} 
& \multicolumn{2}{c}{\textbf{Avg.}} \\
& AUROC & $\text{F}_1$ 
& AUROC & $\text{F}_1$ 
& AUROC & $\text{F}_1$ 
& AUROC & $\text{F}_1$ 
& AUROC & $\text{F}_1$  \\

\midrule

\rowcolor{blue!10}
\multicolumn{11}{c}{\textbf{Training-based Methods}} \\
OpenAI-D        & \large 77.51 & \large 71.18 & \large 78.15 & \large 71.90 & \large 74.60 & \large 70.03 & \large 84.75 & \large 77.47 & \large 78.75 & \large 72.65 \\
Biscope         & \large 79.74 & \large 73.08 & \large 79.97 & \large 73.20 & \large 76.52 & \large 71.64 & \large 92.88 & \large 86.90 & \large 82.28 & \large 76.21 \\
R-Detect        & \large 61.91 & \large 67.14 & \large 67.40 & \large 66.56 & \large 79.19 & \large 73.38 & \large 65.93 & \large 67.72 & \large 68.61 & \large 68.70 \\
\midrule

\rowcolor{blue!10}
\multicolumn{11}{c}{\textbf{Training-free Methods}} \\
Entropy         & \large 83.72 & \large 79.10 & \large 64.30 & \large 71.92 & \large 47.82 & \large 69.24 & \large 75.42 & \large 74.72 & \large 67.82 & \large 73.75 \\
Likelihood      & \large 85.77 & \large 78.38 & \large 66.82 & \large 66.71 & \large 48.96 & \large 66.69 & \large 85.35 & \large 79.75 & \large 71.73 & \large 72.88 \\
LogRank         & \large 87.50 & \large 80.70 & \large 67.30 & \large 66.71 & \large 50.55 & \large 66.69 & \large 86.28 & \large 80.69 & \large 72.91 & \large 73.70 \\
DetectGPT       & \large 73.13 & \large 70.11 & \large 49.57 & \large 66.67 & \large 34.67 & \large 66.67 & \large 78.69 & \large 73.80 & \large 59.02 & \large 69.31 \\
Fast-DetectGPT  & \large 89.77 & \large 84.12 & \large 82.26 & \large 75.93 & \large 74.98 & \large 68.91 & \large 93.25 & \large 90.00 & \large 85.07 & \large 79.74 \\
Binoculars      & \large 90.00 & \large 87.40 & \large 83.21 & \large 82.87 & \large 77.45 & \large 80.20 & \large 93.64 & \large 90.51 & \large 86.08 & \large 85.25 \\
Lastde++        & \large 91.43 & \large 84.97 & \large 75.36 & \large 69.24 & \large 67.30 & \large 66.67 & \large 93.90 & \large 89.41 & \large 82.00 & \large 77.57 \\
IRM & \large 71.85 & \large 70.75  & \large 83.02 & \large 76.46 & \textbf{\large 91.51} & \large 84.05 & \large 77.70 & \large 76.62 & \large 81.02 & \large 76.97 \\
DNA-DetectLLM & \underline{\large 91.74} & \underline{\large 87.72} & \underline{\large 88.97} & \underline{\large 84.85} & \large 88.23 & \underline{\large 84.94} & \underline{\large 94.48} & \underline{\large 90.58} & \underline{\large 90.86} & \underline{\large 87.02} \\
\textbf{Exons-Detect} & \textbf{\large 92.43} & \textbf{\large 88.05} & \textbf{\large 90.67} & \textbf{\large 84.95} & \underline{\large 90.46} & \textbf{\large 86.59} & \textbf{\large 94.98} & \textbf{\large 91.30} & \textbf{\large 92.14} & \textbf{\large 87.72}\\
\bottomrule
\end{tabular}
}
\caption{Detection performance (AUROC and F1 score) on public benchmark datasets.}
\label{tab: main_result}
\end{table*}

\subsection{Main Results}
Table~\ref{tab: main_result} compares the detection performance of Exons-Detect against other baselines across three public benchmarks. Overall, Exons-Detect exhibits strong detection accuracy and robust generalization, delivering consistently competitive results on M4, DetectRL, and RealDet. It achieves an average AUROC of 92.14\% and an average F1 score of 87.72\%, outperforming the latest baseline DNA-DetectLLM by 1.4\% and 0.8\%. Notably, Exons-Detect is the only method whose AUROC exceeds 90\% on all evaluated datasets, further highlighting its reliability for real-world deployment under different text distributions.

A closer inspection reveals that most baselines exhibit substantial performance variance across datasets, indicating sensitivity to changes in text source and distribution. For training-based detectors, the mismatch between the training corpus and the evaluation benchmarks leads to limited OOD generalization, with AUROC typically remaining below 80\%. Among training-free approaches, representative methods such as Fast-DetectGPT, Binoculars, and Lastde++ perform strongly on M4 and RealDet, yet their performance degrades sharply on the more challenging DetectRL setting. IRM excels on DetectRL, reaching an AUROC of 91.51\% under the Multi-Domain setting, but fails to maintain comparable performance on M4 and RealDet. We conjecture that this behavior arises from IRM’s reliance on probability discrepancies induced by RLHF, which become weaker and harder to exploit when texts are generated by less strongly aligned open-source LLMs. While DNA-DetectLLM partially alleviates this issue, it still falls noticeably behind Exons-Detect on DetectRL. In particular, Exons-Detect achieves AUROC gains of 1.9\% under Multi-LLM and 2.5\% under Multi-Domain. We attribute this consistent advantage to Exons-Detect’s ability to reliably identify exonic tokens from hidden-state discrepancies across diverse text distributions, enabling it to extract more precise source-relevant signals and thereby counteract the cross-dataset bias.

\subsection{Robustness}

\begin{figure*}[t]
\centering
\counterwithout{figure}{section}
\resizebox{\linewidth}{!}{
\includegraphics{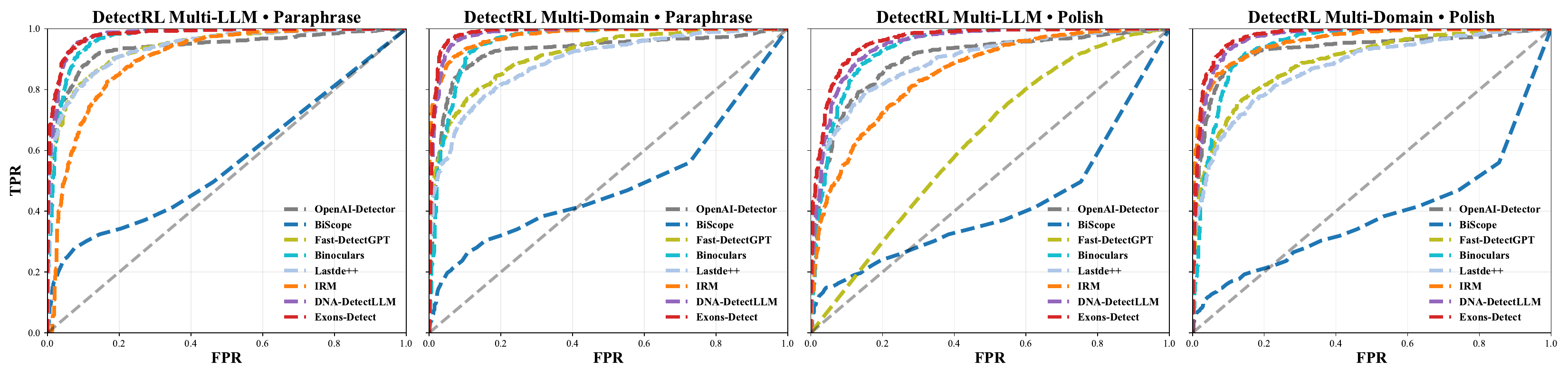}
}
\caption{Detection performance (AUROC curves) against various attacks.
}
\label{fig:robost_attack}
\vspace{-6pt}
\end{figure*}

\begin{figure}[t]
\centering
\counterwithout{figure}{section}
\resizebox{\linewidth}{!}{
\includegraphics{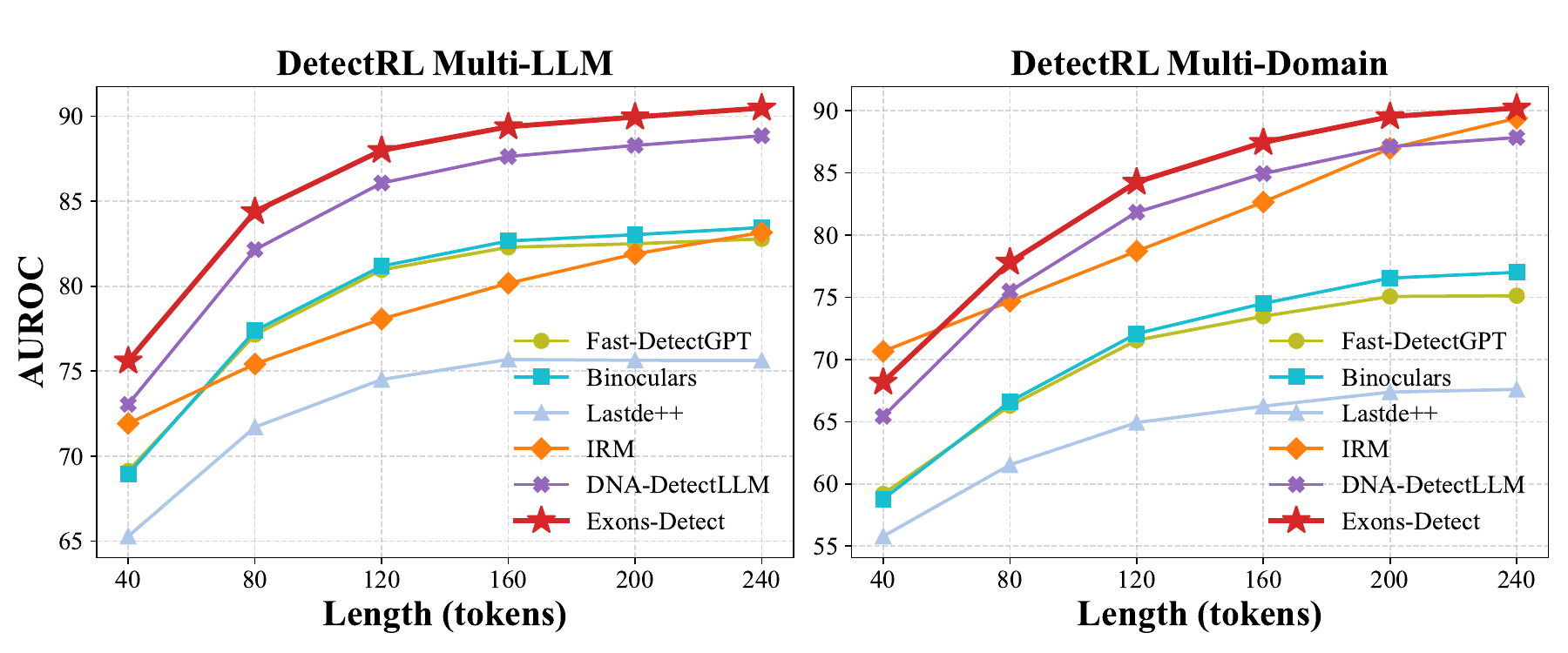}
}
\caption{Detection performance on different length.
}
\label{fig:robost_length}
\end{figure}

\subsubsection{Robustness against Various Attacks}
In realistic scenarios, input texts are often non-pristine and may be subject to adversarial attacks. AI-generated texts can undergo paraphrasing attacks to evade detection, while human-written texts are frequently refined using advanced LLMs through polishing attacks. Paraphrasing attacks employ DIPPER~\cite{NEURIPS2023_575c4500} to rephrase AI-generated texts, while polish attacks apply GPT-4o-based polishing to human-written texts. Figure~\ref{fig:robost_attack} shows the AUROC of Exons-Detect and baselines on DetectRL under various attack settings.

Experimental results demonstrate that Exons-Detect exhibits strong robustness across these attack scenarios. Specially, under both paraphrasing and polishing attacks on DetectRL, Exons-Detect consistently outperforms all competing baselines, maintaining a clear performance advantage. A notable observation is that training-based detectors are substantially more vulnerable to adversarial attacks, as evidenced by BiScope’s AUROC degrading to near-random performance under paraphrasing and polishing attacks. In contrast, strong training-free baselines exhibit substantially higher robustness to paraphrasing attacks, likely because DIPPER-based paraphrasing mainly introduces lexical and syntactic variations without fundamentally altering underlying statistical features. 

However, polishing attacks pose a greater challenge by injecting advanced LLM alignment and generation signals into human-written texts, blurring the boundary between human-written and AI-generated content. This leads to noticeable performance degradation for several baselines, including Fast-DetectGPT and IRM. In contrast, Exons-Detect preserves a high level of detection accuracy even under polishing attacks, highlighting its superior robustness. We attribute this robustness to its ability to exploit hidden-state discrepancies to identify critical exonic tokens, thereby suppressing the adverse impact of localized textual modifications on the global detection score.

\subsubsection{Robustness on Different Lengths}
Prior studies \cite{bao2024fastdetectgpt, tian2024multiscale} have demonstrated that detection performance is highly sensitive to input length, with shorter texts being substantially more difficult to identify. To systematically examine this effect, we truncate input texts to several predefined lengths and evaluate method robustness under varying length constraints. Figure~\ref{fig:robost_length} compares the robustness of Exons-Detect against five strong baselines on DetectRL across different input lengths. The results show that Exons-Detect consistently outperforms all competing baselines across predefined lengths, achieving an average improvement of 2.7\% over DNA-DetectLLM and 6.4\% over IRM. While all methods benefit from increased text length, Exons-Detect exhibits markedly stronger performance in the short-text regime. These results highlight that Exons-Detect captures precise source-related signals from limited text, leading to superior robustness under short-length conditions.

\begin{figure*}[!htbp]
\centering
\counterwithout{figure}{section}
\resizebox{\linewidth}{!}{
\includegraphics{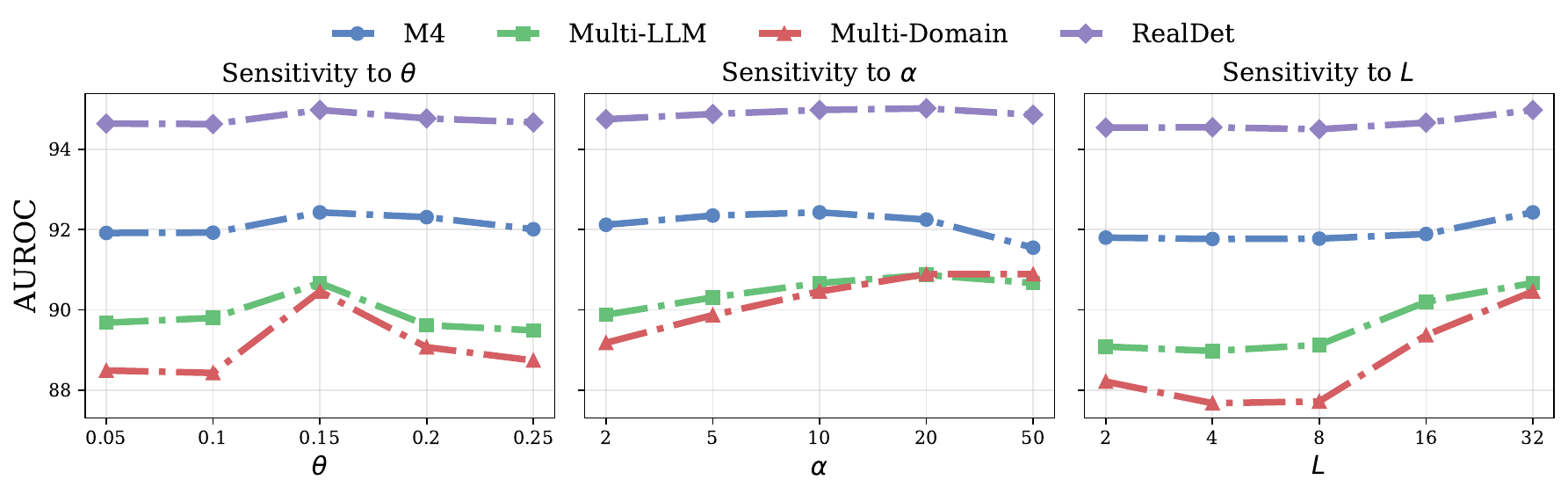}
}
\caption{Detection performance of Exons-Detect under different parameter settings.}
\label{fig:hyperparameter}
\vspace{-6pt}
\end{figure*}

\begin{figure}[t]
\centering
\counterwithout{figure}{section}
\resizebox{\linewidth}{!}{
\includegraphics{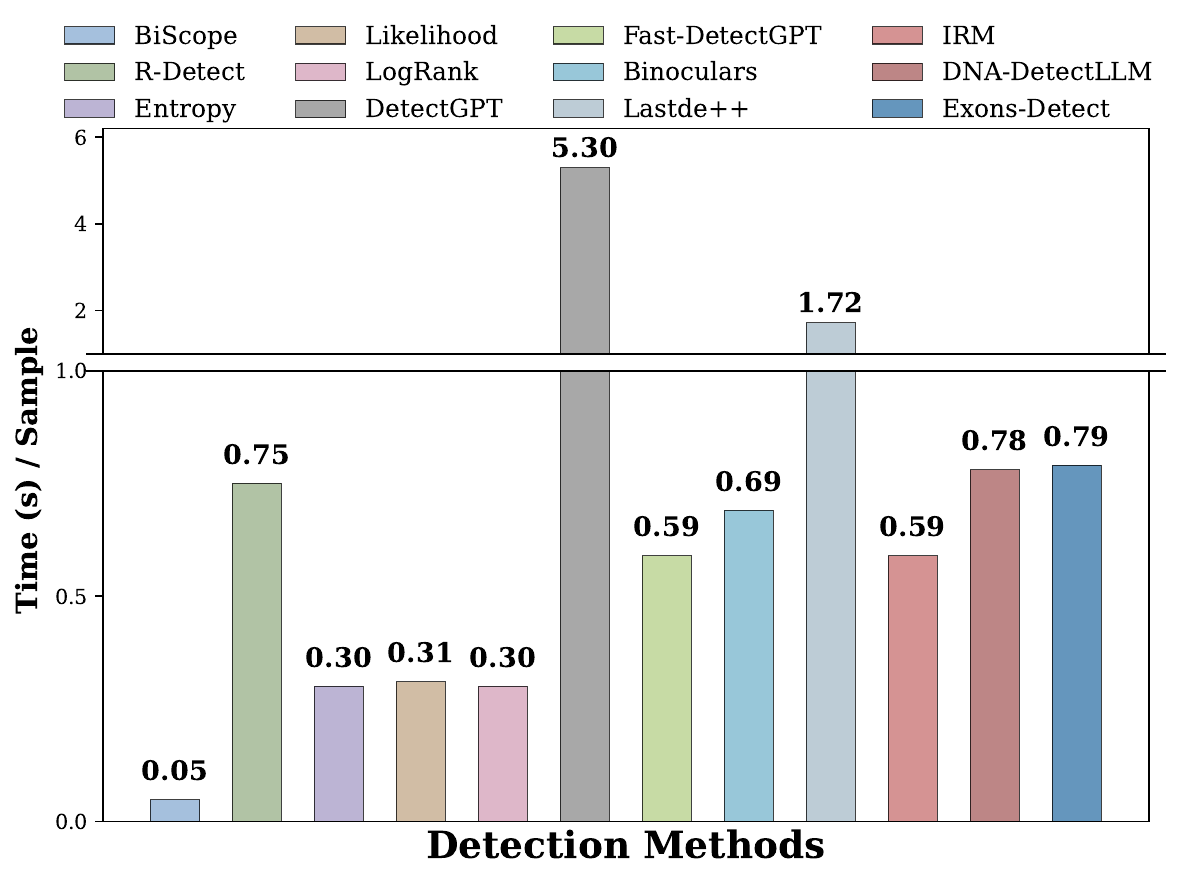}
}
\caption{Time costs for processing a single sample.
}
\label{fig:effiency}
\end{figure}

\subsection{Ablation Studies}
\begin{table}[t]
\centering
\resizebox{\linewidth}{!}{
\begin{tabular}{l|ccccc}
\toprule
\textbf{Setting} $\downarrow$
& \textbf{M4} 
& \textbf{Multi-L} 
& \textbf{Multi-D} 
& \textbf{RealDet} 
& \textbf{Avg.} \\
\midrule

Exons-Detect & \textbf{92.43} & \textbf{90.67} & \textbf{90.46} & \textbf{94.98} & \textbf{92.14} \\
w/o $\log \mathrm{PPL}^W_{M}(\hat{s})$ & 91.28 & 85.32 & 80.46 & 93.97 & 87.76 \\
w/o $g(\cdot)$ & 91.92 & 89.63 & 88.66 & 94.64 & 91.21 \\

\midrule

\textbf{Model Family} $\downarrow$\\

Falcon-7B
& \textbf{92.43} & 90.67 & 90.46 & \textbf{94.98} & \textbf{92.14} \\

LLaMA-7B 
&  88.64 & 92.34 & 90.31 & 94.47 & 91.44 \\
Mistral-v0.1-7B
& 90.41 & 91.42 & 85.77 & 93.00 & 90.15 \\
LLaMA-3.2-1B 
& 90.50 & \textbf{92.87} & \textbf{90.82} & 92.58 & 91.69 \\

\bottomrule
\end{tabular}
}
\caption{Ablation study results under different settings.}
\label{tab:ablation}
\end{table}

\textbf{Impact of the $g(\cdot)$ and  $\log \mathrm{PPL}^W_{M}(\hat{s})$}. 
Table~\ref{tab:ablation} evaluates the impact of removing the nonlinear mapping function $g(\cdot)$ or the computation of $\log \mathrm{PPL}^W_{M}(\hat{s})$ (i.e., the mutation-repair mechanism). Removing $g(\cdot)$ corresponds to assigning additional weights as $\Delta w_t={\delta}_t$, while removing $\log \mathrm{PPL}^W_{M}(\hat{s})$ refers to performing detection using the initial translation score. Overall, both $g(\cdot)$ and $\log \mathrm{PPL}^W_{M}(\hat{s})$ make essential contributions to detection performance and are indispensable components of Exons-Detect. Specifically, removing $g(\cdot)$ and $\log \mathrm{PPL}^W_{M}(\hat{s})$ results in average AUROC drops of 1.0\% and 4.4\%. These degradations indicate that properly mapping hidden-state discrepancies to token importance weights, as well as leveraging the mutation-repair mechanism to further capture class-discriminative differences, are effective and necessary for achieving strong performance.

\noindent\textbf{Impact of the proxy LLM pair.} Table~\ref{tab:ablation} also reports Exons-Detect’s performance across different LLM pairings, including Falcon-7B-Instruct with Falcon-7B, LLaMA-2-7B with LLaMA-7B, Mistral-v0.1-7B-Instruct with Mistral-v0.1-7B, and LLaMA-3.2-1B-Instruct with LLaMA-3.2-1B. Overall, Exons-Detect achieves consistently strong performance across all model combinations, with only modest variation and an average AUROC exceeding 90\% in every setting. Notably, the “LLaMA-3.2-1B-Instruct + LLaMA-3.2-1B” pairing slightly outperforms “Falcon-7B-Instruct + Falcon-7B” on DetectRL, attaining AUROC scores of 92.87\% and 90.82\%. These results indicate that while certain LLM combinations can offer incremental gains, the effectiveness of Exons-Detect does not hinge on a specific model pairing. Instead, the method remains robust across diverse LLM families, and can be further enhanced by selecting better LLM pairings.

\subsection{Hyperparameter Sensitivity}

This subsection analyzes the impact of both hyperparameters (discrepancy threshold $\theta$ and mapping slope $\alpha$) and a structural parameter (hidden layers $L$) on detection performance. Figure~\ref{fig:hyperparameter} reports the detection performance of Exons-Detect under different parameter settings across multiple datasets.

\textbf{Overall, Exons-Detect exhibits low sensitivity to hyperparameter choices, maintaining stable detection performance across a broad range of settings.} Specifically, varying either $\theta$ or $\alpha$ results in performance fluctuations typically within 1.0\%, while consistently outperforming existing baselines under all configurations. For the threshold $\theta$, we observe that extreme values lead to inferior performance compared to values around $\theta = 0.15$. This behavior is intuitive: overly small thresholds tend to treat most tokens as exonic tokens, excessively amplifying noise, whereas overly large thresholds fail to emphasize informative tokens, diminishing the benefit of reweighting. For the mapping slope $\alpha$, performance improves noticeably when $\alpha \geq 10$, indicating that sufficiently steep mappings are necessary to translate moderate hidden-state discrepancies into effective importance weights.

\textbf{Reducing $L$ leads to a clear degradation in detection performance.} For instance, on DetectRL under the Multi-LLM and Multi-Domain settings, reducing $L$ from 32 to 4 results in relative AUROC drops of 1.9\% and 3.1\%. Across all datasets, Exons-Detect consistently achieves its best performance when utilizing 32 hidden layers. These results highlight that fully exploiting representational discrepancies across the entire depth of the model is crucial for robust detection.

\subsection{Efficiency Analysis}
Faster detection is critical for real-world deployment and monitoring. Figure~\ref{fig:effiency} compares the per-text runtime of all methods. To control for the effect of text length, we sample 1,000 long samples from RealDet and truncate each to 300 tokens, reporting the average processing time per text.
Training-based detectors (e.g., BiScope) achieve the lowest inference latency, but at the cost of substantial training overhead. Among training-free methods, classical detectors such as Likelihood are relatively fast (around 0.3 s per text) since they require only a single forward pass, but their detection accuracy did not meet our requirements. Exons-Detect and representative baselines (e.g. Binoculars) incur two forward passes but still run within 0.8 s. Within this efficiency regime, Exons-Detect delivers better detection performance, offering a favorable accuracy-latency trade-off.

\section{Conclusion}
This paper proposes Exons-Detect, a novel training-free method for AI-generated text detection that operates by identifying and reweighting exonic tokens. Extensive experiments demonstrate that Exons-Detect consistently achieves SOTA performance across diverse evaluation settings, while exhibiting strong robustness to adversarial attacks and varying input lengths. We hope our work offers new insights for AI-generated text detection and plan to further explore token-level contribution modeling to enhance detection performance.

\section*{Limitations}
Prior studies \cite{chen2025repreguarddetectingllmgeneratedtext} have shown that hidden-state representations may carry signals related to text provenance. In Exons-Detect, we employ cosine distance to efficiently measure token-level hidden-state discrepancies for assessing token importance. We believe that more fine-grained and more specific discrepancy evaluations could better exploit source-related information and lead to more accurate detection results, which represents a potential direction for further improvement from a token-level perspective.




\bibliography{custom, anthology}

\appendix

\section{Effect Analysis of Exonic Reweighting}
\label{app:exon-proof}

\paragraph{Notation.}
Given a sequence $s=\{x_1,\ldots,x_T\}$, we define the token-level quantities
\begin{equation}
a_i \triangleq -\log P_M(x_i),
\end{equation}
\begin{equation}
b_i \triangleq -\,P_M(x_i)\log P_{\tilde M}(x_i),
\end{equation}
and the unweighted global score
\begin{equation}
R_0 \triangleq \frac{A_0}{B_0}, \qquad
A_0 = \sum_{i=1}^T a_i,\quad
B_0 = \sum_{i=1}^T b_i.
\label{eq:R0_app}
\end{equation}

Let $S=\{i:\delta_i>\theta\}$ denote the set of exonic tokens.
For each $i\in S$, we assign an additional, token-specific weight $\Delta w_i>0$ mapped from the hidden-state discrepancy $\delta_i$;
tokens outside $S$ retain unit weight.
We define the corresponding weighted sums over exonic tokens as
\begin{equation}
A_S \triangleq \sum_{i\in S} \Delta w_i a_i,\qquad
B_S \triangleq \sum_{i\in S} \Delta w_i b_i .
\label{eq:AS_BS_app}
\end{equation}
The resulting exon-aware translation score is then given by
\begin{equation}
R^{W} \triangleq \frac{A_0 + A_S}{B_0 + B_S}.
\label{eq:RW_app}
\end{equation}

\paragraph{Score shift under exon-aware reweighting.}
Since $b_i>0$ for all tokens, it follows directly that $B_0>0$ and $B_S>0$.
We analyze the difference between the reweighted and unweighted scores:
\begin{equation}
\begin{aligned}
R^{W}-R_0
&= \frac{A_0+A_S}{B_0+B_S}-\frac{A_0}{B_0} \\
&= \frac{B_0(A_0+A_S)-A_0(B_0+B_S)}{B_0(B_0+B_S)} \\
&= \frac{B_0A_S-A_0B_S}{B_0(B_0+B_S)} \\
&= \frac{B_0B_S}{B_0(B_0+B_S)}
   \left(\frac{A_S}{B_S}-\frac{A_0}{B_0}\right).
\end{aligned}
\label{eq:diff_app}
\end{equation}
As the denominator is strictly positive, we obtain the following sign equivalence:
\begin{equation}
\mathrm{sign}(R^{W}-R_0)
=
\mathrm{sign}\!\left(\frac{A_S}{B_S}-R_0\right).
\label{eq:sign_app}
\end{equation}

\paragraph{Connection to empirical observations.}
Figure~\ref{fig:method} shows that among tokens with large hidden-state discrepancy
($\delta_i>\theta$), the number of tokens that increase class separability
substantially exceeds the number that decrease it.
In addition, the magnitudes of the corresponding token-level quantities $a_i$ and $b_i$
are empirically observed to be of comparable scale, rather than differing by orders of magnitude.
Taken together, these observations indicate that the aggregated exonic ratio
\begin{equation}
\frac{A_S}{B_S}
=
\frac{\sum_{i\in S} \Delta w_i a_i}{\sum_{i\in S} \Delta w_i b_i}
\end{equation}
is predominantly influenced by tokens that contribute in a label-consistent direction.
Accordingly, for near-boundary (hard) samples with $R_0\approx\tau$, we expect
$\frac{A_S}{B_S}<\tau$ for AI-generated texts and $\frac{A_S}{B_S}>\tau$ for human-written texts.
By~\eqref{eq:sign_app}, this implies $R^{W}-R_0<0$ for AI-generated texts and
$R^{W}-R_0>0$ for human-written texts, indicating that exon-aware reweighting
pushes the score toward the correct side of the decision boundary and improves separability.

\section{Additional Implementation Details}
\label{appendix:implementation}
Regarding the construction of the evaluation datasets, we randomly and uniformly sample 2,000 text samples from each public benchmark, including M4, DetectRL (Multi-LLM and Multi-Domain settings), and RealDet, ensuring balanced class distributions for experimental evaluation.

During evaluation, the maximum input length is capped at 1024 tokens.
All experiments are conducted on a single NVIDIA A100 GPU with 80GB memory.
All models are executed using 32-bit floating-point precision (FP32).

\section{Data Construction in the Robustness Experiment}
In the Polish Attack, we employ GPT-4o to refine texts originally written by humans. The specific model version and decoding parameters are as follows:

\begin{itemize}
\item \textbf{GPT-4o Turbo}: \texttt{gpt-4o-2024-11-20}, Temperature = 1.0, Top-$p$ = 1.0.
\end{itemize}

To ensure that the semantic content and overall structure of the original human-written texts remain largely unchanged, the model is instructed to perform light polishing only, focusing on improving fluency and expression rather than rewriting or altering meaning. The input prompt is carefully constructed to enforce this constraint and is specified as follows:

\begin{itemize}
    \item \texttt{
    Polish the following human-written text by correcting grammar and improving fluency, while ensuring that the semantic content, author intent, and discourse structure remain unchanged. The result should read more natural but convey exactly the same meaning as the original text: \\
    \textbackslash n + original human-written text
    }
\end{itemize}

\begin{table*}[t]
\centering
\resizebox{\textwidth}{!}{
\begin{tabular}{l|cc|cc|cc|cc|cc}
\toprule
\multirow{2}{*}{\textbf{Detectors}} 
& \multicolumn{2}{c|}{\textbf{M4}} 
& \multicolumn{2}{c|}{\makecell{\textbf{DetectRL}  \textbf{Multi-LLM}}} 
& \multicolumn{2}{c|}{\makecell{\textbf{DetectRL}  \textbf{Multi-Domain}}} 
& \multicolumn{2}{c|}{\textbf{RealDet}} 
& \multicolumn{2}{c}{\textbf{Avg.}} \\
& AUROC & $\text{F}_1$ 
& AUROC & $\text{F}_1$ 
& AUROC & $\text{F}_1$ 
& AUROC & $\text{F}_1$ 
& AUROC & $\text{F}_1$  \\

\midrule

\rowcolor{gray!10}
\multicolumn{11}{c}{\textbf{Nonlinear mapping}} \\
$\alpha=10$        & \large 92.43 & \large 88.05 & \large 90.67 & \large 84.95 & \large 90.46 & \large 86.59 & \large 94.98 & \large 91.30 & \large 92.14 & \large 87.72 \\
$\alpha=20$          & \large 92.25 & \large 87.55 & \large 90.87 & \large 85.20 & \large 90.89 & \large 87.05 & \large 95.02 & \large 91.29 & \large 92.26 & \large 87.77 \\

\midrule
\rowcolor{gray!10}
\multicolumn{11}{c}{\textbf{Linear mapping}} \\
$\alpha=10$        & \large 92.24 & \large 87.92 & \large 90.90 & \large 84.93 & \large 91.18 & \large 87.24 & \large 95.03 & \large 91.02 & \large 92.34 & \large 87.78 \\
$\alpha=20$         & \large 91.65 & \large 86.63 & \large 90.66 & \large 84.40 & \large 91.68 & \large 86.85 & \large 95.04 & \large 90.66 & \large 92.26 & \large 87.14 \\

\bottomrule
\end{tabular}
}
\caption{Detection performance (AUROC and F1 score) with different mapping functions ($\theta=0.15$).}
\label{tab: mapping_function}
\end{table*}

\begin{table*}[ht]
\centering
\resizebox{\linewidth}{!}{
\begin{tabular}{l|cc|cc|cc}
\toprule
\multirow{2}{*}{\textbf{Parameter Setting}} 
& \multicolumn{2}{c|}{\makecell{\textbf{DetectRL}  \textbf{Multi-LLM}}} 
& \multicolumn{2}{c|}{\makecell{\textbf{DetectRL}  \textbf{Multi-Domain}}} 
& \multicolumn{2}{c}{\textbf{Avg.}} \\
& AUROC & $\text{F}_1$ 
& AUROC & $\text{F}_1$ 
& AUROC & $\text{F}_1$  \\
\midrule
\rowcolor{gray!10}
\multicolumn{7}{c}{\textbf{Reverse}} \\
$L=2\quad(\theta=0.15, \alpha=10)$ & 89.09 & 84.39 & 88.24 & 84.53 & 88.67 & 84.46 \\
$L=4\quad(\theta=0.15, \alpha=10)$ & 89.41 & 85.07 & 88.38 & 84.65 & 88.90 & 84.86 \\
$L=8\quad(\theta=0.15, \alpha=10)$ & 89.13 & 84.36 & 88.56 & 84.70 & 88.85 & 84.53 \\
$L=16\quad(\theta=0.15, \alpha=10)$ & 90.63 & 85.38 & 90.55 & 86.79 & 90.59 & 86.09 \\
\midrule
\rowcolor{gray!10}
\multicolumn{7}{c}{\textbf{Forward}} \\
$L=2\quad(\theta=0.15, \alpha=10)$ & 89.09 & 84.39 & 88.21 & 84.41 & 88.65 & 84.40 \\
$L=4\quad(\theta=0.15, \alpha=10)$ & 88.98 & 84.38 & 87.68 & 84.12 & 88.33 & 84.25 \\
$L=8\quad(\theta=0.15, \alpha=10)$ & 89.13 & 84.36 & 87.72 & 84.21 & 88.43 & 84.29 \\
$L=16\quad(\theta=0.15, \alpha=10)$ & 90.20 & 85.08 & 89.37 & 85.52 & 89.79 & 85.30 \\
\bottomrule
\end{tabular}
}
\caption{Detection performance with different hidden layers.}
\label{tab:hidden_layer}
\end{table*}

\section{Further Exploration of Mapping Functions}

This section further investigates the impact of different mapping functions on detection performance, focusing on a comparison between a linear mapping and the default nonlinear mapping. The nonlinear mapping is computed as defined in Eq~\ref{Eq:mapping}, while the linear mapping is formalized as follows:
\begin{equation}
g(\delta_t)
=
\alpha \, (\delta_t - \theta)_+.
\label{Eq:linear_mapping}
\end{equation}
Under the linear mapping, the additional weight $\Delta w_t$ is constrained within the range $(0,\alpha)$, which results in a steeper scaling behavior compared to the $(0,1)$ range adopted by the nonlinear mapping. As reported in Table~\ref{tab: mapping_function}, the experimental results show that the linear mapping can still achieve competitive performance. In some cases, it even attains a higher average performance than the nonlinear counterpart. However, its performance tends to be less stable across different datasets.

Overall, these results consistently demonstrate the effectiveness of mapping hidden-state discrepancies to additional token-level weights. Importantly, this effectiveness remains robust to the specific choice of mapping function, indicating that the proposed reweighting mechanism is not sensitive to the particular form of the mapping employed.

\section{Effect of Hidden-Layer Discrepancies}

In the hyperparameter sensitivity analysis, we compared the impact of extracting hidden-layer discrepancies across varying numbers of layers, ranging from 2 to 32, on detection performance. Building upon this analysis, this section further investigates the effect of reverse extraction of hidden-layer discrepancies, specifically considering differences computed from the last layers backward, spanning from 2 to 16 layers.

As reported in Table~\ref{tab:hidden_layer}, the experimental results show that, for the same number of layers, reverse extraction consistently outperforms forward extraction. This observation suggests that discrepancies derived from higher hidden layers are more informative, as they more effectively capture token-level importance at the current position. Moreover, these findings indicate the potential benefits of employing more fine-grained and structurally richer strategies for modeling hidden-layer discrepancies, which may further enhance detection performance.

\begin{table}[t]
\centering
\resizebox{\linewidth}{!}{
\begin{tabular}{l|cccc}
\toprule
\textbf{Setting ($L=16$)} $\downarrow$
& $\theta=0.05$ 
& $\theta=0.10$ 
& $\theta=0.15$ 
& $\theta=0.20$ \\
\midrule
\rowcolor{gray!10}
\multicolumn{5}{c}{\textbf{M4}} \\
$\alpha=2$ & 92.03 & 92.09 & 92.23 & 92.09  \\
$\alpha=6$ & 91.99 & 92.06 & 92.57 & 92.38  \\
$\alpha=10$ & 91.89 & 91.88 & \textbf{92.60} & 92.50  \\

\midrule
\rowcolor{gray!10}
\multicolumn{5}{c}{\textbf{Multi-LLM}} \\
$\alpha=2$ & 89.58 & 89.64 & 89.72 & 89.47 \\
$\alpha=6$ & 89.59 & 89.67 & 90.07 & 89.54  \\
$\alpha=10$ & 89.52 & 89.56 & \textbf{90.20} & 89.60  \\

\midrule
\rowcolor{gray!10}
\multicolumn{5}{c}{\textbf{Multi-Domain}} \\
$\alpha=2$ & 88.57 & 88.66 & 88.89 & 88.64  \\
$\alpha=6$ & 88.42 & 88.49 & 89.27 & 88.87  \\
$\alpha=10$ & 88.26 & 88.18 & \textbf{89.37} & 88.97  \\

\midrule
\rowcolor{gray!10}
\multicolumn{5}{c}{\textbf{RealDet}} \\
$\alpha=2$ & 94.71 & 94.73 & 94.78 & 94.68  \\
$\alpha=6$ & 94.72 & 94.79 & 95.01 & 94.81  \\
$\alpha=10$ & 94.68 & 94.75 & \textbf{95.10} & 94.88  \\
\bottomrule
\end{tabular}
}
\caption{AUROC under different hyperparameters.}
\label{tab:additional_hyper}
\end{table}

\section{Additional Hyperparameter Experiments}

We conduct additional experiments to further examine the effects of the hyperparameters $\alpha$ and $\theta$. Specifically, on public benchmark datasets, we evaluate detection performance under $L = 16$ hidden layers, with $\alpha \in \{2, 6, 10\}$ and $\theta \in \{0.05, 0.10, 0.15, 0.20\}$, as summarized in Table~\ref{tab:additional_hyper}.

It is evident that, regardless of the hyperparameter configuration, utilizing hidden-layer discrepancies from only 16 layers consistently underperforms the setting with 32 layers. Nevertheless, under the 16-layer configuration, \textsc{Exons-Detect} remains largely insensitive to variations in both $\alpha$ and $\theta$, exhibiting only marginal performance fluctuations. Furthermore, an analysis of AUROC score variations indicates that, for the LLM pair (Falcon-7B Instruct + Falcon-7B), the optimal hyperparameter combination is $\alpha = 10$ and $\theta = 0.15$.


\end{document}